\documentclass{article}

    \PassOptionsToPackage{numbers, compress}{natbib}

\usepackage{floatrow}

    \usepackage[final]{neurips_2020}

\usepackage[utf8]{inputenc} 
\usepackage[T1]{fontenc}    
\usepackage{hyperref}       
\usepackage{url}            
\usepackage{booktabs}       
\usepackage{amsfonts}       
\usepackage{nicefrac}       
\usepackage{microtype}      

\usepackage{hyperref}       
\usepackage{amsthm}         
\usepackage{amsmath}        
\usepackage[capitalise]{cleveref}       
\usepackage[ruled,vlined]{algorithm2e} 
\usepackage[dvipsnames]{xcolor} 
\usepackage{kotex}          
\usepackage{graphicx}       
\usepackage{wrapfig}        
\usepackage{multirow}       
\usepackage{multicol}       
\usepackage{tabularx}       

\usepackage{caption}
\usepackage{subcaption}

\title{Pretraining Neural Architecture Search Controllers with Locality-based Self-Supervised Learning}

\author{%

  Kwanghee~Choi\thanks{Equal contributions.} \\
  Hyperconnect \\
  \texttt{kwanghee.choi@hpcnt.com} \\
   \And
  Minyoung~Choe\footnotemark[1] \\
  Sogang University \\
  \texttt{min05@sogang.ac.kr} \\
   \And
  Hyelee~Lee\footnotemark[1] \\
  Naver \\
  \texttt{hyelee.lee@navercorp.com}
}

\graphicspath{{figs/}}

\begin{document}

\maketitle

\begin{abstract}
Neural architecture search (NAS) has fostered various fields of machine learning. Despite its prominent dedications, many have criticized the intrinsic limitations of high computational cost. We aim to ameliorate this by proposing a pretraining scheme that can be generally applied to controller-based NAS. Our method, locality-based self-supervised classification task, leverages the structural similarity of network architectures to obtain good architecture representations. We incorporate our method into neural architecture optimization (NAO) to analyze the pretrained embeddings and its effectiveness and highlight that adding metric learning loss brings a favorable impact on NAS. Our code is available at \url{https://github.com/Multi-Objective-NAS/self-supervised-nas}.
\end{abstract}

\section{Introduction}
Neural architecture search (NAS) has been successful in multiple domains, such as computer vision \cite{zoph2018learning, tan2019efficientnet} or natural language processing \cite{so2019evolved}. NAS optimizes the wiring and operation types of the neural network \cite{xie2019exploring} so that the resulting network can perform well on various target tasks such as object detection \cite{ghiasi2019fpn} or semantic segmentation \cite{liu2019auto}. Typically, NAS methods such as \cite{zoph2016neural, zoph2018learning, luo2018neural, baker2016designing} depend on the controller to generate candidate architectures. Pairs of network architecture and its task performance are fed to the controller and the controller tries to infer better architecture by capturing the underlying relationship between the two. In other words, the controller learns architecture embeddings with respect to the performance of each network.

However, relying solely on the architecture-performance pairs to train the controller has certain drawbacks. First, producing sample pairs requires a tremendous amount of computational cost \cite{zoph2018learning}. Also, structural similarities of the architecture are not considered, where it can be useful to learn better architecture embeddings. For example, isomorphic graphs must have the identical performance, and similar graphs tend to show comparable performance for the target task \cite{ying2019bench}.

We mitigate the aforementioned problems by pretraining the controller on our self-supervised task, \textit{locality-based classification}. We choose a random graph from the target NAS search space as an anchor and generate positive samples on the fly by conducting a few random edits to the graph. Then, metric learning algorithms are applied to embed graphs with low edit-distance in close proximity. Our method does not require expensive architecture-performance pairs, and it can be generally employed to many controller-based NAS algorithms.

Our contributions can be summarized as follows:
\begin{itemize}
    \item We propose a general pretraining strategy for NAS which exploits structural similarities.
    \item We inspect the NAS-Bench-101 \cite{ying2019bench} search space and reformulate the strategy above as an efficient self-supervised classification task.
    \item We incorporate our method to Neural Architecture Optimization \cite{luo2018neural} with various metric learning methods and show the efficacy on NAS.
\end{itemize}

\section{Related Works}
\paragraph{Boosting NAS efficiency}
To overcome the problem of obtaining architecture-performance pairs, several approaches have been suggested. \cite{zoph2018learning} reduces the search space by deconstructing the network architecture into a combination of few smaller motifs. One-shot methods \cite{brock2017smash, pham2018efficient} estimate the fully-trained network by weight sharing. \cite{baker2017accelerating, wen2019neural} predict network performances and \cite{luo2020semi, tang2020semi} adapt the semi-supervised approach to quickly find the best architecture among the search space. These methods, including ours, can be generally used on top of the other.

\paragraph{Exploiting graph locality}
The idea of exploiting similarities between architectures is not new. For example, local search algorithm \cite{ottelander2020local}, network morphism \cite{elsken2017simple} or evolutionary algorithm \cite{real2019regularized, liu2017hierarchical, miikkulainen2019evolving, real2017large, yang2020cars} for NAS assume that similar network structures lead to similar performance. However, they utilize the structural locality only during training, whereas our method does on pretraining.

\paragraph{Embedding network architectures}
To acquire effective representations of network architectures, multilayer perceptrons or recurrent neural networks \cite{hochreiter1997long} are often used to map graphs onto the continuous space \cite{zoph2016neural, liu2018progressive}. Several encoding schemes such as adjacency matrix encoding \cite{zoph2018learning, ying2019bench, wen2019neural} or path encoding \cite{white2019bananas, wei2020npenas} have been proposed to maintain the characteristics of graphs. Also, \cite{wen2019neural} adopt graph convolutional networks \cite{bruna2013spectral, micheli2009neural}. \cite{yan2020does} introduce isomorphism networks \cite{xu2018powerful} and \cite{ning2020generic} present a novel embedding scheme so that the encoder directly captures the graph information. Most similar approach to our work is \cite{cheng2020nasgem}, where similarity loss between cosine distance of embeddings and the Weisfeiler-Lehman kernel \cite{shervashidze2011weisfeiler} value between graphs are suggested.

\paragraph{Metric Learning}
With the rise of deep neural networks, metric learning has been used to acquire discriminant representations in the embedding space by learning distance function over individual samples. There are two general approaches, i.e., training with pair-wise labels and class-level labels. The former directly learns the similarity with positive and negative samples \cite{hadsell2006dimensionality, chopra2005learning, hoffer2015deep, schroff2015facenet, oh2016deep, sohn2016improved, ustinova2016learning, wang2017deep, wu2017sampling, wang2019multi}, and the latter with the classification loss function \cite{ranjan2017l2, liu2017sphereface,liu2016large, wang2017normface, wang2018additive, wang2018cosface, deng2019arcface}.

\begin{figure}%
    \vspace*{-0.5cm}
    \begin{subfigure}{.2\textwidth}
        \centering
        \begin{tabular}{l|l}
             & Ratio \\ \hline
            1 & 0.0002 \\
            2 & 0.0001 \\
            3 & 0.0024 \\
            4 & 0.0073 \\
            5 & 0.0291 \\
            6 & 0.0798 \\
            > 6 & 0.8811 \\
        \end{tabular}
        \\[1em]
        \caption{}
        \label{fig-edit-dist-distrb}
    \end{subfigure}
    \begin{subfigure}{.3\textwidth}
        \centering
        \includegraphics[height=3.8cm]{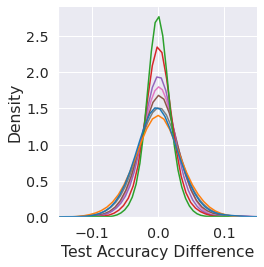}
        \caption{}
        \label{fig-compare-edit-dist}
    \end{subfigure}
    \begin{subfigure}{.4\textwidth}
        \centering
        \includegraphics[height=3.8cm]{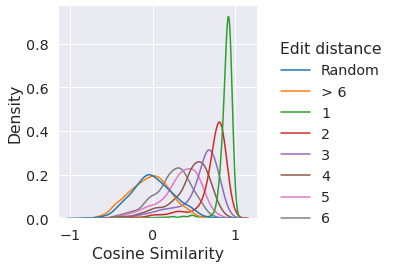}
        \caption{}
        \label{fig-edit-dist-vs-embed-dist}
    \end{subfigure}
    \caption{
        (a) Edit-distance distribution of $10000$ randomly sampled architecture pairs.
        (b) Performance difference distribution of $1000$ randomly sampled architecture pairs per edit-distance.
        (c) Correlation between the edit-distance and the cosine similarity of architecture embeddings which is learned by AngularLoss \cite{wang2017deep}. We plot (b) and (c) using kernel density estimation.
    }
\end{figure}

\section{Locality-based Classification}
We first show that edit-distance between graphs correlates with task performance to design a self-supervised dataset. Then we illustrate how we construct the dataset by generating positive and negative samples on the fly. Finally, to demonstrate the applicability of our self-supervised task, we employ our pretraining scheme on Neural Architecture Optimization \cite{luo2018neural} (NAO) as an example.

\paragraph{Performance locality}
In \cref{fig-compare-edit-dist}, we verify the positive correlation between the edit distance and the performance difference by extending the works of NAS-Bench-101 \cite{ying2019bench}. They focus on the locality near the global optimum architecture, whereas we inspect the entire search space.

\paragraph{Constructing the self-supervised dataset}
To remove the burden of calculating the edit-distance which is compute-intensive, we formulate the edit-distance prediction task into the classification task between edit-distance $\leq 6$ (Positive samples) and $> 6$ (Negative samples). We chose $6$ based on the observations by \cite{ying2019bench}. Positive samples are inexpensively generated by randomly editing nodes and edges from the initial graph. Also, it is safe to assume that the majority of the randomly chosen graph will fall inside $> 6$, as shown in \cref{fig-edit-dist-distrb}. Note that different anchors may share the same positive samples. This helps samples to fill up the manifold uniformly as they maintain moderate distance between each other. This renders a smoother embedding space as we intended.

\paragraph{Applying our method to Neural Architecture Optimization}
NAO \cite{luo2018neural} propose a controller consisting of encoder $f_e$, performance predictor $f_p$, decoder $f_d$ with a jointly trainable loss function
\begin{align}
    \mathcal{L}_{train}=\lambda \mathcal{L}_{p} + (1-\lambda)\mathcal{L}_{r} = \lambda(f_p(f_e(a))-r)^2+(1-\lambda)\log P_d(a|f_e(a))
\end{align}
where $a \in \mathcal{A}$ is the target architecture in the search space $\mathcal{A}$, $r$ is the performance, and $\lambda \in [0,1]$ is the trade-off parameter between the performance prediction loss $\mathcal{L}_{p}$ and the reconstruction loss $\mathcal{L}_{r}$. To find the embedding $e'$ of the new candidate architecture, NAO performs gradient ascent on $f_p$ from the embedding $e=f_e(a_{seed})$ of the previously best architecture $a_{seed}$. Then $f_d$ decodes $e'$ to obtain the next candidate. As the train proceeds, the number of sample pairs increases and both architecture embedding and performance prediction jointly improve.

With our self-supervised task, we pretrain the $f_e$ and $f_d$ with the following loss function. Note that pair-wise label losses are used for $\mathcal{L}_e$, since it is infeasible to train with class-level labels.
\begin{align}
    \mathcal{L}_{pretrain}=\mathcal{L}_{r} + \lambda_e \mathcal{L}_{e} = \mathcal{L}_{r} + \lambda_e \max(d(f_e(a), f_e(p)) - d(f_e(a), f_e(n)) + m, 0)
\end{align}
where $\lambda_e > 0$ is the trade-off parameter and $\mathcal{L}_e$ is the pair-wise metric learning loss for architecture embeddings. TripletMarginLoss \cite{hermans2017defense} is given as an example above, where $m$ is the margin, $d$ is a distance function for the embedding space $f_e(\mathcal{A})$, positive sample architecture $p \in \mathcal{G}(a)$ with graph modifier $\mathcal{G}$ and negative sample architecture $n \in \mathcal{A}$.

\begin{figure}
    \vspace*{-0.5cm}
    \centering
    \begin{subfigure}{.24\textwidth}
        \centering
        \includegraphics[height=3cm]{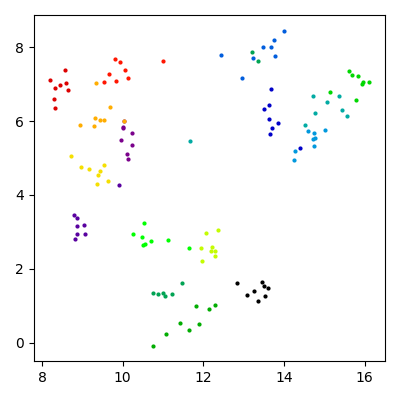}
        \caption{Pretrained MarginLoss}
        \label{fig-umap-pretrained-margin}
    \end{subfigure}
    \begin{subfigure}{.23\textwidth}
        \centering
        \includegraphics[height=3cm]{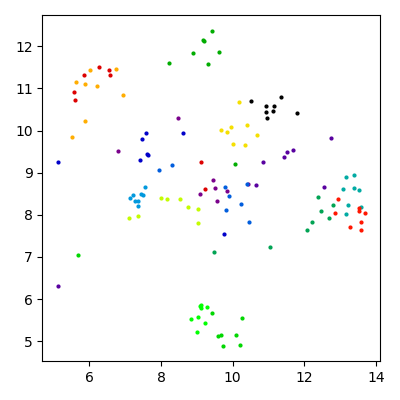}
        \caption{Trained MarginLoss}
        \label{fig-umap-trained-margin}
    \end{subfigure}
    \begin{subfigure}{.26\textwidth}
        \centering
        \includegraphics[height=3cm]{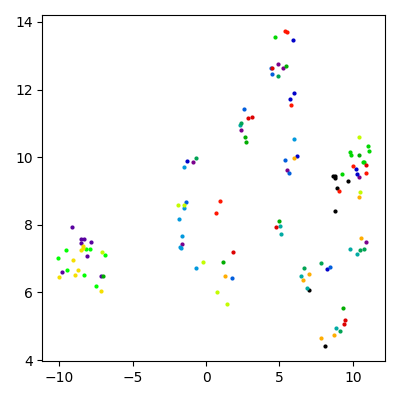}
        \caption{Pretrained NoMetricLoss}
        \label{fig-umap-pretrained-nometric}
    \end{subfigure}
    \begin{subfigure}{.24\textwidth}
        \centering
        \includegraphics[height=3cm]{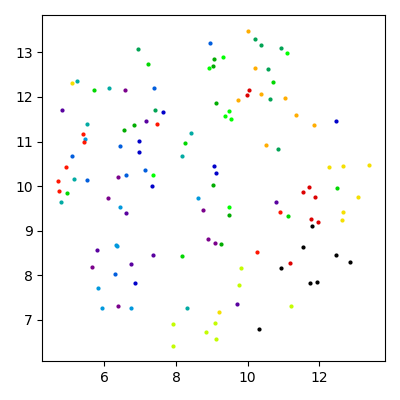}
        \caption{Trained NoMetricLoss}
        \label{fig-umap-trained-nometric}
    \end{subfigure}
    \caption{
        Visualization of network embeddings after fully pretrained or fully trained. We randomly choose $16$ anchor graphs, generating $7$ positive graphs per anchor. Same class samples have the same color. UMAP \cite{mcinnes2018umap} with euclidean distance metric, minimum distance $0.5$ and $32$ neighbors are used.
    }\label{fig-umap}
\end{figure}

\begin{figure}
    \centering
    \subfloat[][Embedding loss $\mathcal{L}_e$]{\includegraphics[height=2.5cm]{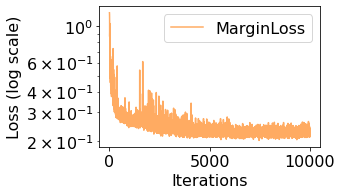}\label{fig-metric-loss}}
    \subfloat[][Reconstruction loss $\mathcal{L}_r$]{\includegraphics[height=2.5cm]{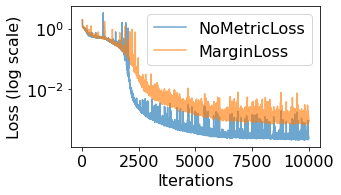}\label{fig-validation-loss}}
    \subfloat[][Reconstruction accuracy]{\includegraphics[height=2.5cm]{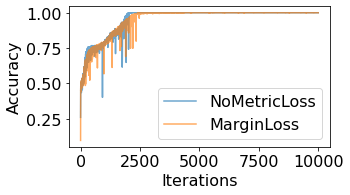}\label{fig-accuracy}}
    \caption{
        Comparison between MarginLoss \cite{wu2017sampling} and NoMetricLoss, both trained for 10k iterations with batch size $512$. Each architecture is decoded as $27$ tokens with 7 possible options each. We define reconstruction accuracy as a classification task with 7 classes for $512 \times 27$ samples per iteration.
    }
\end{figure}

\begin{figure}
    \vspace*{-0.5cm}
    \centering
    \subfloat[\label{fig-better-avg}]{\includegraphics[height=3.5cm]{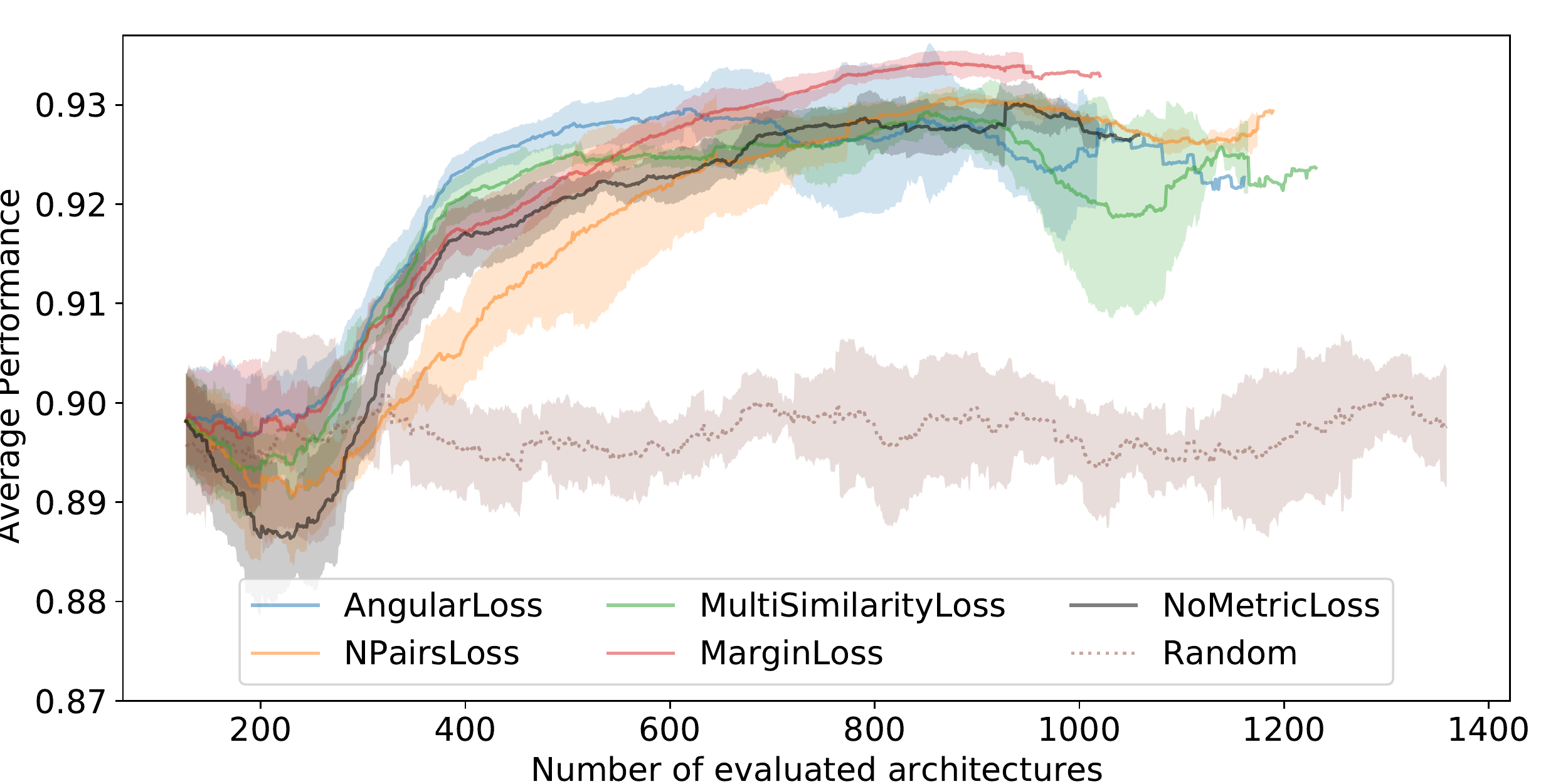}}
    \subfloat[\label{fig-worse-avg}]{\includegraphics[height=3.5cm]{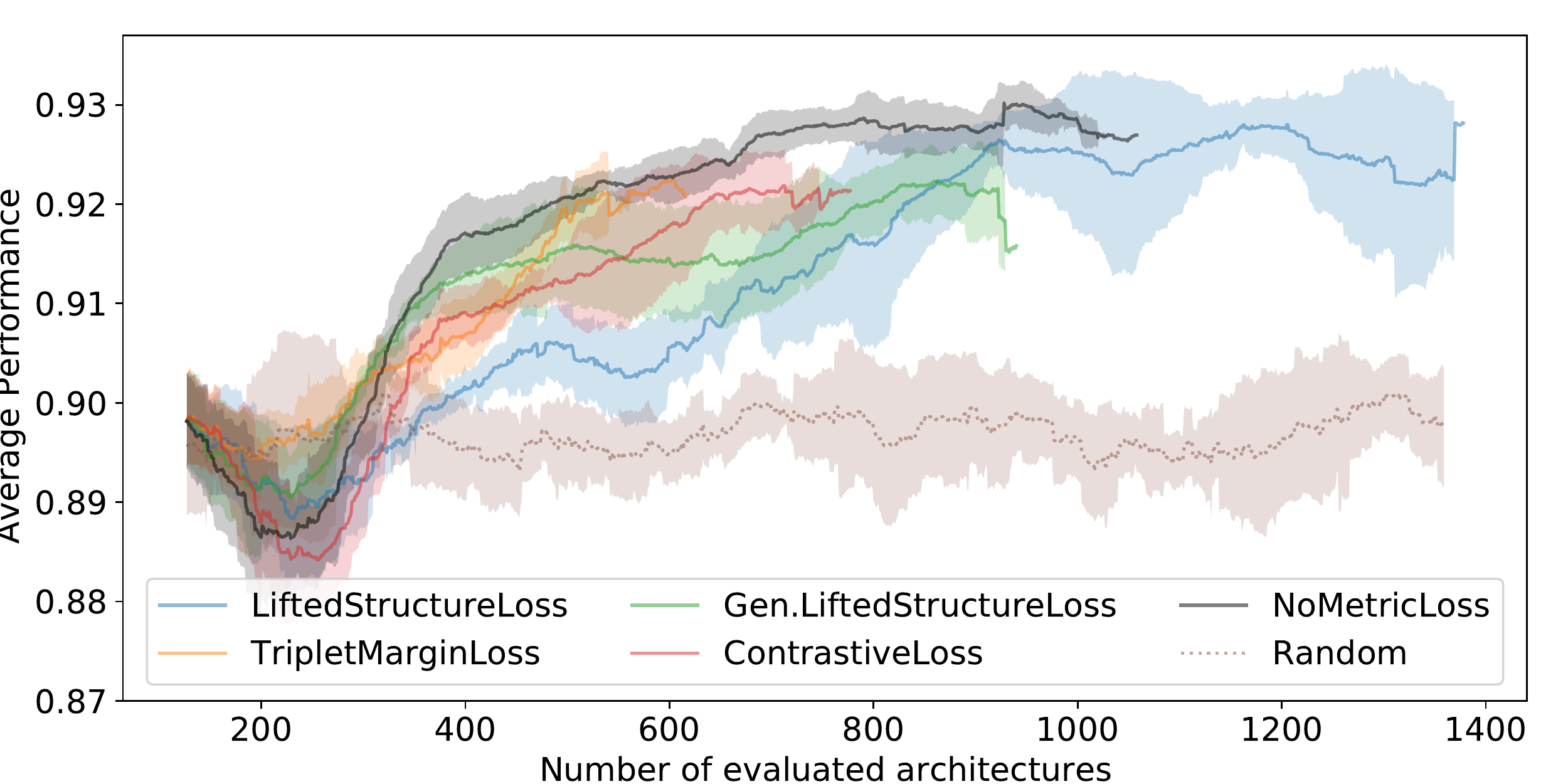}}
    \hfill
    \caption{
        (a), (b) Moving average of the architecture performance with window size $128$. The line is the mean of 5 repetitive runs and the shaded area shows the standard deviation. The number of architectures found are inconsistent among metric learning algorithms. This is because some methods end up generating invalid architectures only.
        Moving average reflects the overall trend of the performance of NAS algorithms.
    }\label{fig-performance}
\end{figure}

\setlength{\extrarowheight}{3pt}
\begin{table}
    \scalebox{0.8}{
    \begin{tabular}{lcccccccccc}
        \toprule
        \cmidrule(r){1-11}
        \multirow{2}{*}{Method} & \multicolumn{2}{c}{\#Queries} & \multicolumn{2}{c}{Test Acc. (\%)} & \multicolumn{2}{c}{SD (\%)} & \multicolumn{2}{c}{Test Regret (\%)} & \multicolumn{2}{c}{Ranking} \\
        \cline{2-11}
        & Mean & Std.. & Mean & Std.. & Mean & Std.. & Mean & Std.. & Mean & Std.. \\ 
        \midrule
        AngularLoss \cite{wang2017deep} & 751 & 141.5 & \bf{94.22} & 0.004 & \bf{3.06} & 1.067 & \bf{0.10} & 0.004 & \textbf{3} & 0.4 \\
        MultiSimilarityLoss \cite{wang2019multi} & 763 & 93.7 & \bf{94.23} & 0.000 & \bf{3.45} & 0.974 & \bf{0.09} & 0.000 & \bf{2} & 0.0 \\
        NPairsLoss \cite{sohn2016improved} & 805 & 188.4 & \bf{94.23} & 0.000 & 3.94 & 1.246 & \bf{0.09} & 0.000 & \bf{2} & 0.0 \\
        \textit{NoMetricLoss} & 810  & 121.2 & \bf{94.23} & 0.000 & 3.94 & 0.797 & \bf{0.09} & 0.000 & \bf{2} & 0.0 \\
        MarginLoss \cite{wu2017sampling} & 838 & 60.7 & \bf{94.23} & 0.000 & \bf{2.80} & 0.936 & \bf{0.09} & 0.000 & \bf{2} & 0.0 \\
        LiftedStructureLoss \cite{oh2016deep} & 1007 & 150.5 & \bf{94.22} & 0.005 & 4.50 & 0.983 & \textbf{0.10} & 0.005 & \textbf{3} & 0.5 \\
        TripletMarginLoss \cite{hermans2017defense} & \bf{519} & 34.5 & 94.07 & 0.138 & 4.27 & 0.860 & 0.25 & 0.138 & 74 & 147.5 \\
        ContrastiveLoss \cite{hadsell2006dimensionality} & 615 & 87.2 & 94.04 & 0.139 & 4.52 & 0.764 & 0.28 & 0.139 & 84 & 118.1 \\
        GenLiftedStructureLoss \cite{hermans2017defense} & 733 & 280.4 & 94.02 & 0.127 & 3.72 & 0.711 & 0.30 & 0.127 & 81 & 94.8 \\
        \textit{Random} & 560 & 248.4 & 93.84 & 0.068 & 6.15 & 1.797 & 0.48 & 0.068 & 336 & 228.1 \\
        
        \bottomrule
    \end{tabular}}
    \caption{Performances of different methods. Baseline methods (\textit{NoMetricLoss} and \textit{Random}) are italicized. "\#Queries" is the number of architecture-accuracy pairs queried until the best performance is found. "SD" is the standard deviation of performances. "Test Regret" is the gap to the best test accuracy in the dataset (94.32\%). "Ranking" is the accuracy ranking of the model found among the whole dataset. Table shows the mean and the standard deviation of 5 repetitive runs.}\label{tab-performance}
\end{table}

\section{Experimental results}

\paragraph{Is edit-distance useful?}
\cref{fig-edit-dist-vs-embed-dist} verifies that architectures with lower edit-distances are embedded closely. Moreover, comparing \cref{fig-umap-pretrained-margin} and \cref{fig-umap-pretrained-nometric} indicates that pretraining with metric loss helps clustering positive samples. Also, clusters are maintained after training (\cref{fig-umap-trained-margin}), and NoMetricLoss starts to identify clusters (\cref{fig-umap-trained-nometric}). These observations imply that the structural similarities act as an informative prior, hence benefiting NAS.

\paragraph{Regularizing nature of $\mathcal{L}_e$}
Adding $\mathcal{L}_e$ results in higher $\mathcal{L}_r$ after convergence (\cref{fig-validation-loss}), but has minimal impact on decoding (\cref{fig-accuracy}). We suspect that $\mathcal{L}_e$ regularizes $\mathcal{L}_r$ by restricting the embedding space to represent the structural similarities. As $\mathcal{L}_e$ pulls all the positive samples closely, decoder gets confused, hence influencing $\mathcal{L}_r$. 

\paragraph{Comparison between metric learning methods}
We compare MarginLoss \cite{wu2017sampling}, AngularLoss \cite{wang2017deep}, NPairsLoss \cite{sohn2016improved}, MultiSimilarityLoss \cite{wang2019multi}, LiftedStructureLoss \cite{oh2016deep}, TripletMarginLoss \cite{hermans2017defense}, ContrastiveLoss \cite{hadsell2006dimensionality} and GenLiftedStructureLoss \cite{hermans2017defense} with the strong baseline \textit{NoMetricLoss}, which is trained only with the reconstruction loss $\mathcal{L}_{r}$, and the weak baseline \textit{Random}, which chooses architectures randomly. $f_e$ and $f_d$ are pretrained with NAS-Bench-101 dataset for $2$ epochs. Pretraining dataset has $7$ positive samples per anchor with $\leq$ $3$ edits, thus the maximum edit-distance is $6$ between positive samples. Training was repeated 5 times to alleviate the impact of randomness. Additional experimental details are attached in Appendix.

\cref{fig-performance} and \cref{tab-performance} illustrate that our method facilitates efficient architecture search.  Besides TripletMarginLoss, ContrastiveLoss and GenLiftedStructureLoss, all find the near-optimal architecture. Furthermore, MultiSimilarityLoss, AngularLoss and NpairLoss take fewer resources than NoMetricLoss, and MultiSimilarityLos, MarginLoss and AngularLoss show less variance than NoMetricLoss. This can be attributed to our pretraining scheme producing well-initialized weights. Especially, MarginLoss and AngularLoss consistently outperform NoMetricLoss in terms of average performance, and MultiSimilarityLoss reaches the optimal architecture the quickest, without suffering from the performance drop. Yet, some losses unexpectedly underperform, failing to find valid architectures as training progresses. The reason for this is unclear, but we suspect that they make the embedding space sparse.

\section{Conclusion}
In this paper, we introduce a novel self-supervised task, \textit{locality-based classification}, which is generally applicable to many controller-based NAS methods. By pretraining architecture embeddings with our self-supervised task, we integrate the structural similarities into the embeddings. Our analysis indicates that structural similarities may act as an informative prior for NAS, providing good initial weights for the controller to boost performance. To further our research, we plan to analyze the impact of regularization and the encoder network structure on the graph embedding space, focusing on why effectiveness differs among metric learning losses.

\section*{Broader Impact}
Our work concentrates on accelerating pre-existing NAS methods. The positive impacts may include reducing computing costs, hence being environment-friendly. However, pushing the frontier of NAS will have mixed consequences. As discussed before, improving meta-algorithms advances deep learning technology in a general sense. Malicious use of the new techniques will naturally follow. Nonetheless, we believe that our work will accelerate research for the benefit of humanity.

\begin{ack}
We thank Yunyoung Choi and Siyeong Lee for valuable discussions, Khu-rai Kim and Younjoon Chung for comments on the paper, and Ui-ryeong Lee for technical assistance.
\end{ack}

\bibliographystyle{plainnat}
{\small
\bibliography{database}
}\clearpage

\section*{Appendix: Implementation Details}
Our implementation is based on PyTorch \cite{paszke2019pytorch} and pytorch-metric-learning \cite{musgrave2020pytorch}. We lightly modified the NAO implementation of SemiNAS \cite{luo2020semi}.

Training was repeated 5 times with different seeds. We run NAO for $50$ epochs (Outer epoch), starting with $128$ random graphs (Initial count) and sampling maximum $128$ graphs (Candidate count) from top $128$ graphs (Seed count) for each epoch with maximum $100$ steps of gradient ascent (Maximum step size). After populating the graph pool, we train for $100$ epochs (Inner epoch). If the controller fails to generate valid architecture or the architecture is already generated in the previous steps, it skips the graph and continues the gradient ascent.

We chose $\lambda_e$ by choosing $\lambda_e=1$. Generally, loss is not sensitive to $\lambda_e$. But, if reconstruction accuracy fails to achieve $1.0$ due to magnitude difference between the reconstruction loss and the embedding loss, we scaled down the embedding loss with the ratio between the converged embedding loss size and the reconstruction loss size. 

\begin{table}[h]
    \caption{Hyperparameters used in pretraining and training}
    \centering
    \begin{tabular}{ll}
    \toprule
    \cmidrule(r){1-2}
        \multicolumn{2}{l}{Pretraining} \\
        \midrule
        Epochs & $2$ \\
        Optimizer & SGD \\
        Learning Rate & $1.0$ \\
        & Decay at iteration $6000$ \\
        & with $\gamma=0.2$ \\
        \bottomrule
   \end{tabular}
  \hfill
   \begin{tabular}{ll}
    \toprule
    \cmidrule(r){1-2}
    \multicolumn{2}{l}{Training} \\
    \midrule
    $\lambda$ & $0.5$ \\
    Gradient clipping & $5.0$ \\
    Batch size & $64$ \\
    Optimizer & SGD \\
    Learning rate & $0.1$ \\
    \bottomrule
    \end{tabular}
\end{table} 

\begin{table}[h]
  \caption{Hyperparameters used for each metric learning loss}
  \centering
  \begin{tabular}{lllll}
    \toprule
    \cmidrule(r){1-5}
    Loss & Loss parameters & Miner &  & Additional parameters\\
    \midrule
    AngularLoss \cite{wang2017deep} & $\alpha=45^{\circ}$ &  Angular-  & Angle $=20^{\circ}$ & Batch size $=256$ \\
    & &Miner \cite{wang2017deep} & & $\lambda_e = 0.1$ \\
 
    \midrule
    Contrastive- & Margin${_+}=0$ & TripletMargin-  & Margin $=0.1$  & Batch size $=512$ \\
    Loss \cite{hadsell2006dimensionality} & Margin$_{-}=0$ & Miner \cite{hermans2017defense} & Semihard & $\lambda_e = 1.0$ \\
 
    \midrule
    GenLifted- & Margin${_+}=0$ & TripletMargin-  & Margin $=0.1$ & Batch size $=512$ \\
    StructureLoss \cite{hermans2017defense} & Margin$_{-}=1$ & Miner \cite{hermans2017defense}& Semihard & $\lambda_e = 0.1$ \\
 
    \midrule
    LiftedStructure- & Margin${_+}=0$ & TripletMargin-  & Margin $=0.1$ & Batch size $=64$ \\
    Loss \cite{oh2016deep}& Margin$_{-}=1$ &Miner \cite{hermans2017defense} & Semihard & $\lambda_e = 0.05$ \\
 
    \midrule
    MarginLoss \cite{wu2017sampling} & Margin $=0.2$ & TripletMargin-  & Margin $=0.2$ & Batch size $=512$ \\
    & $\nu = 0.0$  &Miner \cite{hermans2017defense} & Semihard & $\lambda_e = 1.0$\\
    & Fixed $\beta=1.2$ & & &  \\
    \midrule
    MultiSimilarity-& $\alpha=2$ &  Angular-  & Angle $=20^{\circ}$ & Batch size $=512$ \\
    Loss \cite{wang2019multi} & $\beta=30$ & Miner \cite{wang2017deep}& & $\lambda_e = 1.0$ \\
    & Base$=1.0$ & & &  \\
    \midrule
    NPairsLoss \cite{sohn2016improved} & &  Angular-  & Angle $=20^{\circ}$ & Batch size $=512$ \\
    & &Miner \cite{wang2017deep} & & $\lambda_e = 1.0$ \\
 
    \midrule
    TripletMargin- & Margin$=0.1$ & TripletMargin-  & Margin $=0.1$ & Batch size $=512$ \\
    Loss \cite{hermans2017defense}& & Miner \cite{hermans2017defense}& Semihard & $\lambda_e = 1.0$ \\
 
    \bottomrule
    \end{tabular}
\end{table}

\end{document}